\documentclass[11pt]{article}

\usepackage[final]{acl}

\usepackage{times}
\usepackage{latexsym}
\usepackage{booktabs}
\usepackage{makecell}

\usepackage[T1]{fontenc}

\usepackage[utf8]{inputenc}

\usepackage{microtype}

\usepackage{inconsolata}

\usepackage{graphicx}
\usepackage{amsmath}
\usepackage{multirow}
\usepackage{xcolor}
\usepackage{pifont}
\usepackage[table]{xcolor}

\newcommand{\cmark}{\textcolor{green}{\ding{51}}} 
\newcommand{\xmark}{\textcolor{red}{\ding{55}}}   
\newcommand\blfootnote[1]{%
  \begingroup
  \renewcommand\thefootnote{}\footnote{#1}%
  \addtocounter{footnote}{-1}%
  \endgroup
}

\title{TRACE: Evidence Grounding-Guided Multi-Video Event Understanding and Claim Generation}

\author{
 \textbf{Pengyu Yan\textsuperscript{1,*}},
 \textbf{Akhil Gorugantu\textsuperscript{1,*}},
 \textbf{Mahesh Bhosale\textsuperscript{1}},
 \textbf{Abdul Wasi\textsuperscript{1}},
\\
 \textbf{Vishvesh Trivedi\textsuperscript{2}},
 \textbf{David Doermann\textsuperscript{1}}
\\
\\
 \textsuperscript{1}University at Buffalo, SUNY,
 \textsuperscript{2}New York University
\\
 \small{
   \textbf{Correspondence:} \href{mailto:pyan4@buffalo.edu}{pyan4@buffalo.edu}
 }
}

\begin{document}

\maketitle

\begin{abstract}
Multi-video event understanding demands models that can locate and attribute query-relevant evidence scattered across long, heterogeneous video corpora. Existing large vision–language models (LVLMs) often underperform in this regime because they quickly exhaust their context budget and struggle to precisely localize evidentially important segments, frequently missing dense informational cues such as broadcast graphics, subtitles, and scoreboards.
We introduce TRACE, an evidence grounding-guided framework that follows a \textit{ground-before-reasoning} strategy for multi-video event reasoning. Our approach first builds a structured, text-searchable timeline for each video using OCR and object detection. A text-only LLM then conducts query-aware evidence localization, selecting relevant moments prior to any downstream visual reasoning. The retrieved frames and their grounding summaries are subsequently used to steer LVLM-based claim generation and cross-video citation consolidation.
Experiments on MAGMaR 2026 and WikiVideo demonstrate that structured grounding markedly boosts factual completeness and attribution fidelity. On the MAGMaR validation split, TRACE raises macro-average MiRAGE F1 from 0.705 to 0.811 compared to an unguided Qwen3-VL-30B baseline, with especially strong improvements in citation recall (0.440 $\rightarrow$ 0.628). The method also attains state-of-the-art results on the official MAGMaR 2026 leaderboard.
Code is released at \url{https://github.com/pengyu965/TRACE}.
\end{abstract}


\blfootnote{*Equal Contribution}

\section{Introduction}
\begin{figure*}[t]
  \centering
  \includegraphics[width=\textwidth]{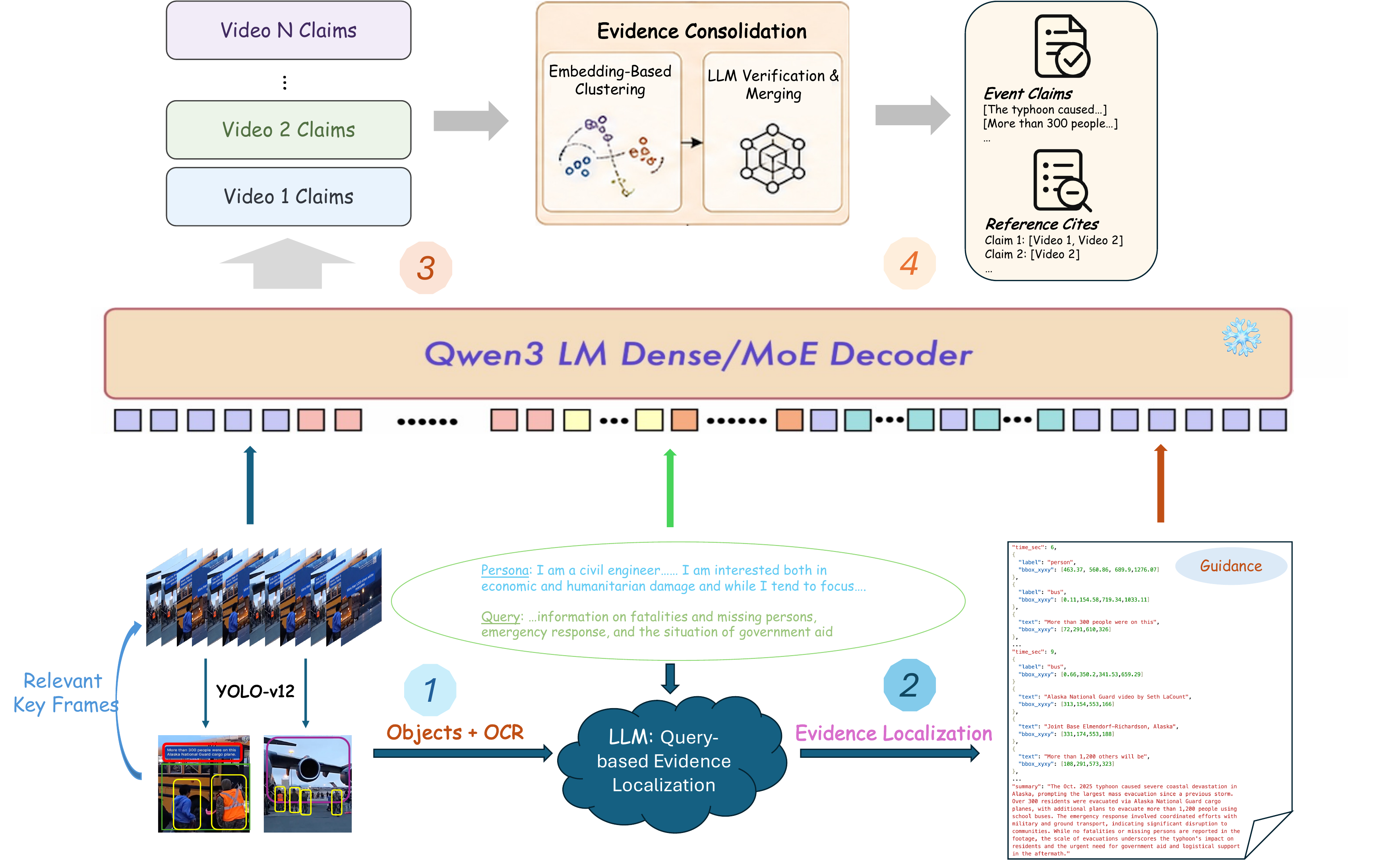}
\caption{
\textbf{Grounding-guided pipeline for event video claim generation.}
We extract structured grounding signals via object detection and OCR over video frames, then use a text-only LLM to align detected labels and on-screen text with the query and persona to identify relevant moments.
This text-based grounding bridges the gap between coarse detector outputs and precise query intent, producing structured guidance that directs the LVLM to relevant timestamps and conditions claim generation on explicit evidence, resulting in factual, well-grounded claims with video citations.
}
\label{fig:pipeline}
\end{figure*}

Multi-video event understanding requires models not only to recognize visual content, but to identify and attribute the specific pieces of evidence that answer a user’s information need. Unlike conventional video captioning, event-centric queries often depend on sparse yet highly informative moments distributed across long collections of heterogeneous videos: a casualty count appearing briefly in a news ticker, a vote total displayed on a broadcast overlay, or an evacuation statistic mentioned alongside supporting footage. Generating factual, grounded claims from such collections is therefore fundamentally an evidence localization problem before it is a generation problem.

Recent large vision--language models (LVLMs) have demonstrated strong capabilities in generic video understanding, yet they remain poorly suited for this setting. When prompted directly with raw video, LVLMs tend to allocate attention toward visually salient content rather than query-relevant evidence, producing broad narrative summaries instead of precise, attributable claims~\citep{martin2025wikivideo}. At the same time, long-video understanding remains constrained by context capacity: even modern LVLMs can process only a limited number of frames, forcing aggressive temporal subsampling that frequently omits the brief moments containing critical information~\citep{wu2024longvideobench,song2024moviechat}. Scaling context windows alone does not resolve this bottleneck, because the challenge is not merely seeing more frames, but identifying which frames matter.

We observe that event videos contain a rich source of lightweight semantic grounding signals that existing LVLM pipelines largely underutilize. Broadcast overlays, captions, scoreboards, banners, and object co-occurrence patterns often encode the exact entities, statistics, locations, and activities required to answer factual queries. In many cases, these structured signals are more semantically informative than the raw visual appearance itself. Crucially, such signals can be extracted efficiently through OCR and object detection without invoking expensive visual reasoning~\citep{hunyuanvisionteam2025hunyuanocrtechnicalreport,tian2025yolov12}.

Motivated by this observation, we propose a \textit{grounding-before-reasoning} paradigm for multi-video event understanding. Instead of asking an LVLM to jointly discover evidence and generate claims from raw video, we first construct a structured, text-searchable representation of each video using OCR and object detection. A text-only LLM then aligns this grounding timeline with the query and persona to identify evidentially relevant moments and synthesize semantic guidance before any visual generation occurs. The downstream LVLM subsequently operates on a targeted subset of frames conditioned on explicit grounding cues, while a final aggregation stage consolidates claims and citations across videos.

This design directly addresses the two central failure modes of current LVLM systems. Query-conditioned grounding concentrates visual capacity on evidentially relevant moments, mitigating context saturation, while structured OCR and detection cues redirect attention toward semantically meaningful content instead of dominant visual patterns. Because the grounding stage is lightweight, text-serializable, and interpretable, it also provides a scalable alternative to brute-force long-context video processing.

We evaluate our approach on the MAGMaR 2026 Oracle Track and WikiVideo benchmarks~\citep{martin2025wikivideo}. Our method achieves state-of-the-art performance on the MAGMaR leaderboard, improving macro-average MiRAGE F1 by 8.2\% over the strongest unguided Qwen3-VL baseline, with especially large gains in citation recall. The same pipeline also generalizes effectively to WikiVideo, demonstrating that lightweight structured grounding transfers across datasets and event domains.

Our contributions are summarized as follows:
\begin{itemize}
    \item We introduce a grounding-guided pipeline that constructs a structured, text-searchable video timeline through OCR and object detection, enabling query-conditioned evidence localization prior to expensive visual reasoning.

    \item We propose a ground-before-reasoning paradigm that separates evidence discovery from multimodal generation, improving both factual completeness and citation attribution in multi-video event understanding.

    \item We design a hybrid grounding and aggregation framework that combines targeted keyframe selection with cross-video claim deduplication and citation propagation.

    \item We achieve state-of-the-art results on the MAGMaR 2026 benchmark and demonstrate strong generalization on WikiVideo, with particularly large improvements in citation recall.
\end{itemize}

\section{Related Work}

\subsection{Multi-Video Event Understanding}

Recent benchmarks have shifted video understanding from generic captioning and QA toward event-centric reasoning over large collections of heterogeneous videos. MultiVENT~\citep{sanders2023multivent,kriz2025multivent} introduces multilingual event retrieval across diverse broadcast and user-generated sources, while WikiVideo~\citep{martin2025wikivideo} studies grounded article generation from multiple event videos. These benchmarks highlight a core challenge of multi-video understanding: relevant evidence is often temporally sparse and distributed across many partially redundant sources. Our work focuses on this evidence localization bottleneck and proposes a grounding-guided framework for query-conditioned claim generation and attribution.

\subsection{Long-Context Video Understanding}

Large vision--language models (LVLMs) such as Video-LLaVA~\citep{lin2024videollava}, VideoChat~\citep{li2024videochat}, and Qwen3-VL~\citep{team2025qwen3} have become the dominant paradigm for video understanding. However, long-video reasoning remains fundamentally constrained by limited visual context capacity. Existing works address this challenge through memory compression~\citep{song2024moviechat}, adaptive frame selection~\citep{tang2025adaptive}, hierarchical representations~\citep{ma2024llamavid}, and sparse temporal sampling~\citep{wu2024longvideobench}. Recent benchmarks including Video-MME~\citep{fu2024videomme} further demonstrate that uniformly sampled frames frequently miss short but information-dense moments critical for downstream reasoning. Our work similarly targets long-context reasoning, but approaches the problem from an evidence-grounding perspective rather than purely improving visual memory or temporal scaling.

\subsection{Query-Guided Localization and Multimodal Grounding}

A large body of work studies grounding language queries to temporally localized video evidence. Temporal localization and moment retrieval approaches such as Moment-DETR~\citep{lei2021momentdetr}, QVHighlights~\citep{lei2021qvhighlights}, and referential video understanding systems~\citep{qiu2024artemis} aim to identify video segments relevant to natural-language queries. In parallel, multimodal grounding approaches including GroundingDINO~\citep{liu2023groundingdino}, GLIP~\citep{li2022glip}, and Kosmos-2~\citep{peng2023kosmos2} align textual semantics with visual entities and regions. Our work differs from these approaches in that we use lightweight OCR and object detections as structured semantic grounding signals for downstream evidence routing and claim generation across multiple videos.

\subsection{OCR and Structured Semantic Signals}

Event videos contain rich structured semantic cues embedded in overlays, captions, scoreboards, and broadcast graphics. OCR-based multimodal reasoning benchmarks such as TextVQA~\citep{singh2019textvqa}, ST-VQA~\citep{biten2019scene}, ChartReformer~\cite{yan2024chartreformer}, and OCR-VQA~\citep{mishra2019ocrvqa} demonstrate the importance of scene text for factual visual understanding. Meanwhile, OCR systems including PaddleOCR~\citep{du2020ppocr} and HunyuanOCR~\citep{tencent2025hunyuanocr} provide efficient extraction of textual evidence from visual content. Similar interactions between graphical structure and embedded text have also been explored in document and chart understanding~\citep{yan2024chartreformer}. Our framework extends these ideas to long-video event understanding, where OCR often provides more semantically precise evidence than raw visual appearance alone.

\subsection{Retrieval-Augmented and Modular Multimodal Reasoning}

Recent multimodal systems increasingly separate evidence discovery from downstream reasoning through retrieval-augmented or modular architectures. Retrieval-augmented language models~\citep{lewis2020rag,borgeaud2022retro,asai2024selfrag} improve factuality by retrieving supporting evidence prior to generation, while modular multimodal systems such as Visual Programming~\citep{gupta2023visual}, ViperGPT~\citep{suris2023vipergpt}, HuggingGPT~\citep{shen2023hugginggpt}, and MM-REACT~\citep{yang2023mmreact} demonstrate the effectiveness of decomposing perception and reasoning into specialized stages. Our work extends this paradigm to multi-video event understanding by introducing a grounding-guided framework that performs lightweight semantic evidence localization before expensive multimodal reasoning.

\section{Method}
\label{sec:method}

Our goal is to generate factual, query-conditioned claims from a collection of event videos while preserving explicit attribution to supporting sources. Rather than relying on an LVLM to jointly discover evidence and perform generation directly from raw video, we decompose the task into two main stages: lightweight evidence grounding followed by grounding-guided multimodal reasoning.

The central idea of our approach is to transform long, unstructured videos into a structured semantic representation that can be efficiently searched and filtered prior to expensive visual inference. We first extract lightweight grounding signals through OCR and object detection to construct a text-searchable timeline of each video. A text-only LLM then performs query-conditioned evidence localization over this timeline, identifying the moments most relevant to the user query and persona. Finally, an LVLM generates claims conditioned on both the selected frames and their associated semantic guidance. Claims from multiple videos are subsequently consolidated through cross-video evidence aggregation. An overview of the pipeline is shown in Figure~\ref{fig:pipeline}.

\subsection{Structured Video Grounding}
\label{sec:method:detection}

Long event videos contain large amounts of redundant visual content interspersed with sparse but highly informative evidence-bearing moments. Processing all frames uniformly with an LVLM is both computationally inefficient and poorly aligned with the evidence localization nature of the task. We therefore first convert each video into a lightweight structured grounding representation that can be queried efficiently before downstream visual reasoning.

\paragraph{Object detection.}
YOLOv12~\cite{tian2025yolov12} processes each sampled frame, yielding per-frame detections
$\mathcal{D}_t = \{(l_i, c_i, \mathbf{b}_i)\}_i$,
where $l_i \in \mathcal{L}_\text{COCO-80}$, $c_i \in [0,1]$, and $\mathbf{b}_i$ is the axis-aligned bounding box.
Object co-occurrence patterns carry rich contextual signal beyond individual labels:
the simultaneous presence of \texttt{person}, \texttt{microphone}, and \texttt{podium}
reliably identifies a press-conference segment without any scene-level supervision.

\paragraph{Text recognition.}
An OCR module extracts visible text strings from each frame.
Broadcast lower-thirds, scoreboards, and graphical overlays name entities, statistics,
and locations that object detectors cannot recover, making on-screen text the highest-precision
signal in news and event footage.

The two streams are merged into a chronological timeline
\begin{equation}
  \mathcal{F} = \bigl\{(t,\, \mathcal{D}_t,\, \mathcal{T}_t)\bigr\}_{t=0}^{T}.
  \label{eq:timeline}
\end{equation}
Because $\mathcal{F}$ is fully text-serializable, the subsequent grounding step is
vision-free — fast, deterministic, and interpretable.

\subsection{Query-Conditioned Evidence Localization}
\label{sec:method:grounding}

The structured grounding timeline provides dense semantic coverage of the video, but only a small subset of frames are typically relevant to a given query and persona. Rather than performing expensive multimodal reasoning over the entire video, we first localize evidentially relevant moments using a lightweight text-only reasoning stage.

\paragraph{Evidence localization.}
A key challenge is that low-level detector outputs do not naturally align with open-ended user queries. Relevant evidence is often expressed indirectly through combinations of OCR text, object co-occurrence, and contextual cues rather than explicit keyword matches. For example, an election-related query may correspond to frames containing vote percentages, podium scenes, and broadcast overlays even when no detector label directly references elections. We therefore introduce a query-conditioned grounding stage that bridges the gap between perception outputs and semantic intent. The timeline $\mathcal{F}$ is partitioned into non-overlapping windows $\{\mathcal{F}_j\}$ of $C$ consecutive frames.
Each window is serialized into a compact textual representation containing timestamps, detected objects, and OCR text. And they are prompted to the LLM alongside $q$ and $p$; the model returns the relevant subset
$\mathcal{S}_j \subseteq \mathcal{F}_j$ together with the supporting detections and OCR strings.
The union
\begin{equation}
  \mathcal{S} = \bigcup_j \mathcal{S}_j
  \label{eq:relevant_frames}
\end{equation}
constitutes the query-relevant keyframe set for that video. Importantly, this stage operates entirely in text space without invoking a vision encoder, making evidence localization substantially more efficient than dense LVLM inference.

\paragraph{Grounding summary.}
Frame-level detections and OCR signals provide sparse semantic anchors, but downstream LVLM generation still requires higher-level contextual understanding of how these observations relate to the query and persona. We therefore introduce an intermediate grounding-summary stage that compresses localized evidence into a coherent semantic description prior to visual generation. This summary acts as a semantic bridge between low-level perception outputs and downstream multimodal reasoning, transforming fragmented detector observations into an interpretable representation of the underlying event narrative.

\subsection{Grounding-Guided Claim Generation}
\label{sec:method:generation}

After evidence localization, the downstream LVLM performs claim generation conditioned on both the original video content and the structured grounding signals. 

\paragraph{Hybrid frame selection.} We construct the LVLM input using a hybrid frame-selection strategy that combines uniformly sampled frames with guidance-targeted evidence frames. 

The visual input to the LVLM is the union
\begin{equation}
  \mathcal{I}_v = \mathcal{I}_{\text{unif}} \;\cup\; \bigl\{\hat{i}_s : t_s \in \mathcal{S}\bigr\},
  \label{eq:frame_set}
\end{equation}
where $\mathcal{I}_{\text{unif}}$ comprises $N_{\text{unif}}$ linearly spaced frames for broad narrative coverage, and each relevant timestamp $t_s$ is mapped to its nearest frame index
\begin{equation}
  \hat{i}_s = \min\!\Bigl(\bigl\lfloor t_s \cdot \mathrm{fps} \rceil,\; F_{\mathrm{total}} - 1\Bigr).
  \label{eq:frame_index}
\end{equation}
After deduplication, frames are sorted temporally and decoded at $448\times448$ pixels.
The uniform sampling preserves broad temporal coverage and guards against potential errors and noise introduced during grounding, while targeted frames allocate visual capacity toward moments identified as evidentially relevant. 

\paragraph{Temporal alignment.}
Frame indices (Eq.~\ref{eq:frame_index}) are passed as explicit positional metadata rather than dense ranks $0, 1, \ldots, N{-}1$.
This preserves correct temporal spacing in the model's rotary position embeddings, letting the LVLM correlate textual grounding annotations (e.g., ``$t{=}45\,\text{s}$: on-screen seat count'') with their visual tokens.
Without this alignment, the text and visual temporal axes diverge, undermining cross-modal grounding.

\paragraph{Evidence fusion.}
The five evidence streams are assembled into a single prompt and passed to the LVLM to generate per-video claims:
\begin{equation}
  \mathcal{C}_v = \mathrm{LVLM}\!\left(\mathcal{I}_v,\; q,\; p,\; \mathcal{A}_{\mathcal{S}},\; g,\; \mathrm{ASR}_v\right),
  \label{eq:generation}
\end{equation}
where $\mathcal{A}_{\mathcal{S}}$ denotes the structured frame-level annotations derived from $\mathcal{S}$ — each entry recording the timestamp, detected objects, and OCR strings of a relevant keyframe.
The remaining inputs are the hybrid frame set $\mathcal{I}_v$, the query $q$ and persona $p$, the grounding summary $g$, and the Whisper ASR transcript $\mathrm{ASR}_v$.
Annotations are cast as \emph{supplementary grounding hints} that the model must cross-validate against the video, preventing over-reliance on potentially noisy detector outputs.
The model is instructed to output single-sentence claims grounded in directly observed evidence, with a preference for specific facts (names, numbers, dates) over vague paraphrases.

\subsection{Cross-Video Claim Consolidation}
\label{sec:method:aggregation}

We frame aggregation as a cross-video evidence consolidation problem rather than a simple textual deduplication task. The goal is not merely to suppress repeated claims, but to reconcile semantically equivalent evidence across videos while preserving the full set of supporting sources.

To achieve this, we first encode generated claims into a semantic embedding space and perform conservative similarity-based clustering. Candidate clusters are subsequently verified by an LLM operating under a strict same-proposition criterion, allowing the system to distinguish genuine paraphrases from superficially similar but factually distinct claims. For each cluster, we retain the most information-complete claim as the canonical representation and propagate the union of supporting video citations across all cluster members. This strategy improves citation recall by explicitly consolidating evidence distributed across multiple videos while avoiding the precision degradation associated with aggressive generative merging.

\begin{table*}[t]
\centering
\caption{\textbf{Official MAGMaR 2026 Leaderboard} (best submission per team, selected entries).
We calculate the F1 and Avg. F1 based on the Info/Cite P/R offered by the MAGMaR 2026 workshop.
Our results leads all teams on all Recall and F1 measures
and ranks second in human evaluation, trailing the top team by only 0.008 points. The baseline model is CAG in~\cite{martin2025wikivideo}}
\small
\setlength{\tabcolsep}{5pt}
\begin{tabular}{lc c c ccc ccc}
\toprule
\multirow{2}{*}{\textbf{Team}} & \multirow{2}{*}{\makecell[c]{\textbf{Human}\\ \textbf{Evaluation}}} & \multirow{2}{*}{\makecell[c]{\textbf{Best}\\\textbf{Votes}}}
& \multirow{2}{*}{\makecell[c]{\textbf{Avg. F1}}}
& \multicolumn{3}{c}{\textbf{Reference Info}}
& \multicolumn{3}{c}{\textbf{Reference Cite}} \\

\cmidrule(lr){5-7} \cmidrule(lr){8-10} 

&  &  & & P & R & F1 & P & R & F1 \\
\midrule
HAIVLab             & 2.526 &  2          & \underline{0.455} & 0.584          &
\underline{0.450} & 0.508          & 0.479          & \underline{0.347} & \underline{0.402}
\\
CiteChasers         & 2.542 &  0          & 0.349             & 0.609          & 0.304
     & 0.406          & 0.509          & 0.204             & 0.291             \\
MARS-Bullet         & 2.667 &  0          & 0.424             & 0.711          & 0.394
     & 0.507          & 0.604          & 0.237             & 0.340             \\
MARS-ss-qa-base     & 3.070 &  6          & 0.299             & 0.331          & 0.306
     & 0.318          & 0.277          & 0.281             & 0.279             \\
Baseline (CAG)      & 3.088 &  1          & 0.434             & \underline{0.764} & 0.410
     & \underline{0.534} & \underline{0.617} & 0.228         & 0.333             \\
MARS-Ginger         & 3.123 &  6          & 0.433             & \textbf{0.776} & 0.404
     & 0.531          & \textbf{0.643} & 0.226             & 0.334             \\
MARS-RLM            & 3.298 &  3          & 0.436             & 0.708          & 0.385
     & 0.499          & 0.592          & 0.272             & 0.373             \\
MARS-iter-qa-ginger & 3.694 &  5          & 0.278             & 0.345          & 0.290
     & 0.315          & 0.257          & 0.226             & 0.241             \\
MARS-ss-qa-ginger   & 3.421 & \textbf{10} & 0.341             & 0.544          & 0.324
     & 0.406          & 0.326          & 0.238             & 0.275             \\
MARS-iter-qa-base   & \textbf{3.833} & \underline{8} & 0.296    & 0.347          & 0.313
       & 0.329          & 0.268          & 0.258             & 0.263             \\
\midrule
\rowcolor{gray!15}
\textbf{Ours} & \underline{3.825} & \underline{8}
& \textbf{0.499}
& 0.640 & \textbf{0.483} & \textbf{0.551}
& 0.498 & \textbf{0.405} & \textbf{0.447} \\
\bottomrule
\end{tabular}
\label{tab:leaderboard}
\end{table*}

\section{Experiments}
\label{sec:experiments}

\subsection{Implementation Details}

\paragraph{Models and hardware.}
All pipeline stages run on four NVIDIA RTX A6000 GPUs (48\,GB each; 192\,GB total VRAM).
The LLM stages — temporal grounding filter and cross-video aggregation — use
\textbf{Qwen3-30B-A3B-Instruct}~\cite{qwen3technicalreport} in BF16 precision,
while the LVLM claim generation stage uses \textbf{Qwen3-VL-30B-A3B-Instruct}~\cite{qwen3technicalreport} in BF16.
Both models are served via vLLM with tensor parallelism
across all four GPUs and are loaded sequentially, so the full 192\,GB budget is
available to each stage.

\paragraph{Token budget.}
Frames are resized to $448\times448$ pixels, yielding approximately 256 visual tokens
per frame under Qwen3-VL's visual tokenizer.
With $N_{\text{unif}}=100$ uniform frames and at most 30 guidance-targeted keyframes,
the visual token ceiling is $130\times256=33{,}280$.
Text context — query, persona, frame annotations, and ASR transcript — contributes
approximately 3{,}600 additional tokens, placing a typical prompt at
${\sim}29{,}000$ tokens, comfortably within the 32{,}768-token context window.

\subsection{Experimental Setup}

\paragraph{Datasets.}
Our primary benchmark is the \textbf{MAGMaR 2026 Oracle Track} validation set,
comprising 8 event topics drawn from real-world news events.
Each topic is paired with a curated set of relevant videos and gold claims
annotated with per-claim video citations.
To assess generalization, we additionally evaluate on the \textbf{WikiVideo} dataset,
which contains 52 queries paired with multi-video collections spanning diverse topics, which is 398 unique videos in total,
using the same pipeline and evaluation protocol.

\paragraph{Evaluation metrics.}
For automatic evaluation, MiRAGE~\cite{martin2025seeingmirageevaluatingmultimodal}
assesses predictions along two axes: \textbf{Reference Info} (InfoP/R), measuring factual completeness of predicted claims
against the gold set, and \textbf{Reference Cite} (CiteP/R), measuring accuracy of per-claim video citations.
Each entailment judgment within MiRAGE is produced by CLUE~\cite{zhang2026unifiedmultimodaluncertaininference}.
We compute F1 scores from the reported precision and recall via the harmonic mean,
and additionally report \textbf{Avg.\ F1}, the macro-average of InfoF1 and CiteF1, as a single summary statistic.
In MAGMaR workshop, human evaluation is conducted from three annotators scoring each system on a 1--5 scale across five dimensions:
factuality, adequacy, coherence, relevancy, and fluency;
they additionally select the single best response per query as vote number.

\begin{table*}[t]
  \centering
  \caption{\textbf{Comparison with LVLM baselines} on the MAGMaR 2026 Oracle Track validation set (8 topics). Our grounding-guided system achieves the highest Avg.\ F1 (0.811), with gains concentrated in citation recall.}
  \small
  \setlength{\tabcolsep}{4pt}
  \begin{tabular}{l | c | ccc | ccc}
  \toprule
  \textbf{Method} &
  \multirow{2}{*}{\makecell[c]{\textbf{Avg.} $\mathbf{F1}$}} &
  \multicolumn{3}{c|}{\textbf{Reference Info}} & \multicolumn{3}{c}{\textbf{Reference Cite}}
  \\
  \cmidrule(lr){3-5} \cmidrule{6-8}
   & & P & R & F1 & P & R & F1 \\
  \midrule

  Qwen3.5-9B   & 0.472 & 0.437 & 0.756 & 0.554 & 0.875 & 0.251 & 0.390 \\
  Qwen3-VL-8B  & 0.723 & 0.870 & \underline{0.802} & \underline{0.835} & 0.93  & \underline{0.452} & 0.608 \\
  Qwen3-VL-30B  & 0.705 & \textbf{0.883} & 0.731 & 0.800 & \textbf{0.990} &
  0.440 & \underline{0.609} \\
  \midrule
  \midrule
  \textbf{Ours} & \textbf{0.811}     &
  \underline{0.863} & \textbf{0.876} & \textbf{0.869} & \underline{0.939} & \textbf{0.628}
  & \textbf{0.753} \\
  \bottomrule
\end{tabular}
\label{tab:results}
\end{table*}

\begin{table}[t]
\centering
\caption{\textbf{Generalization to WikiVideo} (52 queries, 398 videos).
Avg.\ F1 is the macro-average of InfoF1 and CiteF1. Our pipeline maintains the highest Avg.\ F1 and citation recall, consistent with MAGMaR findings.}
\small
\setlength{\tabcolsep}{5pt}
\renewcommand{\arraystretch}{1.15}
\begin{tabular}{l ccc}
\toprule
\textbf{Metric}
& \textbf{Qwen3-VL-8B}
& \textbf{Qwen3-VL-30B}
& \textbf{Ours} \\
\midrule
Avg.\ F1
& \underline{0.878}
& 0.854
& \textbf{0.879} \\
\midrule
\rowcolor{blue!12}
\multicolumn{4}{c}{\textbf{\textit{Reference Info}}} \\
\quad P   & \textbf{0.915}    & \underline{0.888} & 0.868          \\
\quad R   & 0.885             & \underline{0.905} & \textbf{0.918} \\
\quad F1  & \textbf{0.885}    & 0.880             & \underline{0.882} \\
\midrule
\rowcolor{blue!12}
\multicolumn{4}{c}{\textbf{\textit{Reference Cite}}} \\
\quad P   & \underline{0.991} & \textbf{0.993}    & 0.936          \\
\quad R   & \underline{0.792} & 0.767             & \textbf{0.838} \\
\quad F1  & \underline{0.871} & 0.828             & \textbf{0.876} \\
\bottomrule
\end{tabular}
\label{tab:wikivideo_results}
\end{table}

\subsection{Official Workshop Results}
\label{sec:leaderboard}

Table~\ref{tab:leaderboard} presents the official MAGMaR 2026 workshop leaderboard.
Our results achieves the highest scores on most automatic
metrics, and achieves the highest final F1 score: InfoF1 0.551, CiteF1 0.447, and Avg.\ F1 0.499,
exceeding the second-ranked team (HAIVLab) by \textbf{+0.049} in \textbf{Avg.\ F1}.
Notably, our Avg.\ F1 also surpasses the workshop-provided CAG baseline~\cite{martin2025wikivideo} by \textbf{+0.065},
which achieves the second-highest InfoP (0.764) among all teams despite its lower recall.

In human evaluation — where annotators score factuality, adequacy, coherence, relevancy,
and fluency on a 1--5 scale — we rank second with $\underline{3.825}$, trailing
MARS-iter-qa-base by only $0.008$ while matching their tally of 8 ``best'' votes.
Notably, MARS-ss-qa-ginger receives the most best votes (10) despite ranking lower
in both scalar human score and automatic metrics, suggesting that pairwise preference
captures a distinct quality dimension from scalar ratings.
The strong alignment between our automatic metric leadership and near-top human
evaluation provides evidence that the MiRAGE framework is a reliable proxy for
human judgment on this task.

\subsection{Comparison with LVLM Baselines}
\label{sec:results}

\paragraph{Baselines.}
We compare against three LVLM baselines that receive no grounding guidance.
\textbf{Qwen3.5-9B} is a compact vision--language model applied directly to the
video and query--persona context.
\textbf{Qwen3-VL-8B} is a medium-scale VLM baseline using uniform frame sampling only.
\textbf{Qwen3-VL-30B} shares the identical backbone with our pipeline but is prompted
with video frames and query--persona context alone, without object detection, OCR, or any LLM grounding filter, directly isolating the contribution of our
multi-modal grounding stage.

Since ground truth annotations are unavailable for the full workshop test set,
we conduct controlled comparisons on the MAGMaR validation subset (8 topics),
for which gold claims and per-claim citations are provided.
Table~\ref{tab:results} reports MiRAGE scores on this set.
Our grounding-guided pipeline outperforms all baselines across every metric,
with the best configuration achieving Avg.\ F1 of \textbf{0.811} versus
0.705 for the strongest baseline (Qwen3-VL-30B), a gain of $+0.106$.

\paragraph{Citation recall is the primary bottleneck for unguided models.}
Qwen3-VL-30B achieves high citation precision (0.990) but very low recall (0.440):
without grounding, the model anchors on the most salient video while overlooking
the broader evidence base.
Our structured guidance record raises CiteR from $0.440$ to \textbf{0.628}
($+42.7\%$ relative) and CiteF1 from 0.609 to \textbf{0.753}, demonstrating
that the grounding stage directs the model to cite the full range of relevant sources.
Citation precision remains high at 0.939, confirming that the additional citations
are well-grounded rather than spurious.

\paragraph{Factual completeness also improves substantially.}
InfoF1 rises from 0.800 (Qwen3-VL-30B) to \textbf{0.869} under our best
configuration, reflecting that grounding-conditioned generation produces claims
that more thoroughly cover the gold annotation.
This gain is consistent across all four of our variants ($\geq 0.859$),
confirming that it stems from the grounding stage itself rather than from any
particular downstream choice.

\begin{table*}[t]
  \centering
  \caption{\textbf{Ablation study} on the MAGMaR 2026 Oracle Track validation set (8 topics), examining the effect of guided keyframe augmentation and aggregation strategy. Embedding-based aggregation and keyframe augmentation provide complementary gains, with their combination achieving the best overall result.}
  \small
  \setlength{\tabcolsep}{4pt}
  \begin{tabular}{c c | c | ccc | ccc}
  \toprule
   \multirow{2}{*}{\makecell[c]{Additional\\Key Frames}} &
  \multirow{2}{*}{\makecell[c]{Aggregation\\Method}} &
  \multirow{2}{*}{\makecell[c]{\textbf{Avg.} $\mathbf{F1}$}} &
  \multicolumn{3}{c|}{\textbf{Reference Info}} & \multicolumn{3}{c}{\textbf{Reference Cite}}
  \\
  \cmidrule(lr){4-6} \cmidrule{7-9}
     & & & P & R & F1 & P & R & F1 \\
  \midrule
   \xmark & LLM       & 0.802              & 0.860          &
  0.858          & 0.859          & 0.921          & 0.626          & 0.745          \\
  \xmark & Embed-Sim                                & \underline{0.808}  & 0.862          &
   0.873          & \underline{0.868} & 0.925       & \textbf{0.628} & \underline{0.748} \\
   \cmark & LLM                                      & 0.804              & 0.849          &
   \textbf{0.885} & 0.867          & 0.931          & 0.616          & 0.741          \\
   \cmark & Embed-Sim                                & \textbf{0.811}     &
  \underline{0.863} & \underline{0.876} & \textbf{0.869} & \underline{0.939} & \textbf{0.628}
  & \textbf{0.753} \\
  \bottomrule
\end{tabular}
\label{tab:ablation}
\end{table*}

\subsection{Generalization to WikiVideo}
\label{sec:wikivideo}

Table~\ref{tab:wikivideo_results} evaluates the same pipeline on WikiVideo,
a larger and more diverse dataset with 52 queries.
Our method achieves Avg.\ F1 of \textbf{0.879}, edging both Qwen3-VL-8B
($0.878$) and Qwen3-VL-30B ($0.854$).
The pattern of gains mirrors MAGMaR: citation recall improves most
($0.792 \to \mathbf{0.838}$) and CiteF1 remains the highest among all methods
($\mathbf{0.876}$).
The smaller absolute margins on WikiVideo reflect the already-high baseline
performance on this dataset.

We attribute the reduced performance gap to two structural differences between the datasets.
First, WikiVideo videos are considerably shorter (mean $60.1\,\text{s}$, median $55.3\,\text{s}$)
compared to MAGMaR (mean $104.9\,\text{s}$, median $58.4\,\text{s}$), and their duration
distribution is more uniform (std $47.6\,\text{s}$ vs.\ $120.8\,\text{s}$).
For shorter, temporally compact videos, uniform frame sampling already provides dense coverage
of the visual content, diminishing the marginal benefit of our YOLO- and OCR-guided keyframe selection.
Second, unguided baselines already achieve near-ceiling performance on WikiVideo
(Avg.\ F1 $\geq 0.854$), leaving limited headroom for further improvement.
Together, these factors explain why the grounding advantage observed on MAGMaR
($+0.106$ over Qwen3-VL-30B) does not fully transfer to WikiVideo ($+0.025$),
while the consistent citation recall gain ($+0.046$ CiteR) confirms that
multi-modal grounding remains beneficial even in this easier regime.

\subsection{Ablation Study}
\label{sec:ablation}

\paragraph{Our variants.}
We evaluate four configurations of our pipeline varying two dimensions:
(i)~\textbf{Frame selection} — uniform 100 frames only (\xmark) versus
uniform frames augmented with guidance-targeted keyframes (\cmark); and
(ii)~\textbf{Aggregation} — LLM-based cross-video merging (\textsc{LLM}) versus
embedding-similarity deduplication with LLM verification (\textsc{Embed-Sim}).

Table~\ref{tab:ablation} breaks down the contribution of each pipeline component.
All four variants comfortably exceed the strongest baseline
(Avg.\ F1 $\geq 0.802$ vs.\ $0.705$), confirming that multi-modal grounding
is the dominant source of improvement regardless of downstream configuration.

\paragraph{Embedding-based aggregation is consistently better.}
\textsc{Embed-Sim} aggregation outperforms \textsc{LLM} aggregation in both
frame-selection settings ($+0.006$ and $+0.007$ in Avg.\ F1, respectively).
The advantage is most visible in CiteF1 ($0.748$ vs.\ $0.745$; $0.753$ vs.\ $0.741$),
suggesting that similarity-based deduplication is more precise at suppressing
redundant claims than purely generative merging.

\paragraph{Guided keyframe augmentation provides complementary gains.}
Adding guidance-targeted keyframes (\cmark) improves InfoR from $0.858$ to $\mathbf{0.885}$
under \textsc{LLM} aggregation, indicating that the additional frames expose
query-relevant visual evidence missed by uniform sampling.
Gains are modest under \textsc{Embed-Sim} ($+0.003$ Avg.\ F1),
suggesting that the text-based grounding signal already captures much of this
context at the prompt level.
The combination of guided keyframes and \textsc{Embed-Sim} aggregation yields
the best overall result: Avg.\ F1 \textbf{0.811}, InfoF1 \textbf{0.869},
and CiteF1 \textbf{0.753}.

\section{Conclusion}
\label{sec:conclusion}

We presented a grounding-guided pipeline for multi-video event claim generation that adopts a ground-then-generate paradigm: lightweight detection and OCR signals direct a text-only LLM to query-relevant keyframes before any visual inference, and a downstream LVLM generates attributed claims conditioned on the resulting guidance.
The approach consistently outperforms unguided LVLM baselines on both MAGMaR 2026 and WikiVideo, with the largest gains in citation recall — confirming that structured perception-based grounding is an effective and transferable principle for video claim attribution.

\paragraph{Limitations and future work.}
The pipeline's object detector is constrained to the COCO-80 vocabulary, limiting its ability to identify domain-specific entities central to many news queries.
The sequential, non-differentiable design also means grounding errors propagate without recovery.
Future directions include open-vocabulary detection~\citep{liu2023groundingdino}, adaptive frame sampling for fast-paced events, timestamp-level citation attribution, and end-to-end joint optimization of the grounding and generation stages.

\newpage
\bibliography{custom}



\end{document}